\pdfoutput=1

\documentclass[11pt]{article}

\usepackage[preprint]{acl}

\usepackage{times}
\usepackage{latexsym}

\usepackage[T1]{fontenc}

\usepackage[utf8]{inputenc}

\usepackage{microtype}

\usepackage{inconsolata}

\usepackage{graphicx}

\usepackage{amsmath} 
\usepackage{amsfonts}  
\usepackage{booktabs, multirow} 
\usepackage{subcaption} 
\usepackage{xcolor, colortbl} 
\usepackage{amssymb}
\usepackage{pifont}
\usepackage[most]{tcolorbox}
\usepackage{soul}
\usepackage{changepage,threeparttable} 

\newcommand{\cmark}{\textcolor{green!80!black}{\checkmark}} 
\newcommand{\xmark}{\textcolor{red}{\ding{55}}} 

\tcbset{
  aibox/.style={
    top=10pt,
    colback=white,
    colframe=black,
    colbacktitle=black,
    enhanced,
    center,
    attach boxed title to top left={yshift=-0.1in,xshift=0.15in},
    boxed title style={boxrule=0pt,colframe=white,},
    fontupper=\small,
  }
}
\newtcolorbox{AIbox}[2][]{aibox, title=#2,#1}

\title{ConvSearch-R1: Enhancing Query Reformulation for Conversational Search with Reasoning via Reinforcement Learning}

\author{
 \textbf{Changtai Zhu\textsuperscript{1}}\thanks{Work done during internship at ByteDance Inc.},
 \textbf{Siyin Wang\textsuperscript{1}},
 \textbf{Ruijun Feng\textsuperscript{3}},
 \textbf{Kai Song\textsuperscript{2}},
 \textbf{Xipeng Qiu\textsuperscript{1}}\thanks{Corresponding author.}
\\ 
 \textsuperscript{1}Fudan University,
 \textsuperscript{2}ByteDance Inc,
 \textsuperscript{3}University of New South Wales
\\ 
\texttt{ctzhu23@m.fudan.edu.cn, \{siyinwang20, xpqiu\}@fudan.edu.cn}, \\
\texttt{songkai.sk.batman@bytedance.com, ruijun.feng@unsw.edu.au}
}

\begin{document}
\maketitle
\begin{abstract}
Conversational search systems require effective handling of context-dependent queries that often contain ambiguity, omission, and coreference. Conversational Query Reformulation (CQR) addresses this challenge by transforming these queries into self-contained forms suitable for off-the-shelf retrievers. 
However, existing CQR approaches suffer from two critical constraints: high dependency on costly external supervision from human annotations or large language models, and insufficient alignment between the rewriting model and downstream retrievers. 
We present ConvSearch-R1, the first self-driven framework that completely eliminates dependency on external rewrite supervision by leveraging reinforcement learning to optimize reformulation directly through retrieval signals. Our novel two-stage approach combines Self-Driven Policy Warm-Up to address the cold-start problem through retrieval-guided self-distillation, followed by Retrieval-Guided Reinforcement Learning with a specially designed rank-incentive reward shaping mechanism that addresses the sparsity issue in conventional retrieval metrics. 
Extensive experiments on TopiOCQA and QReCC datasets demonstrate that ConvSearch-R1 significantly outperforms previous state-of-the-art methods, achieving over 10\% improvement on the challenging TopiOCQA dataset while using smaller 3B parameter models without any external supervision.
\end{abstract}

\section{Introduction}
\begin{figure}[t]
  \includegraphics[width=\columnwidth]{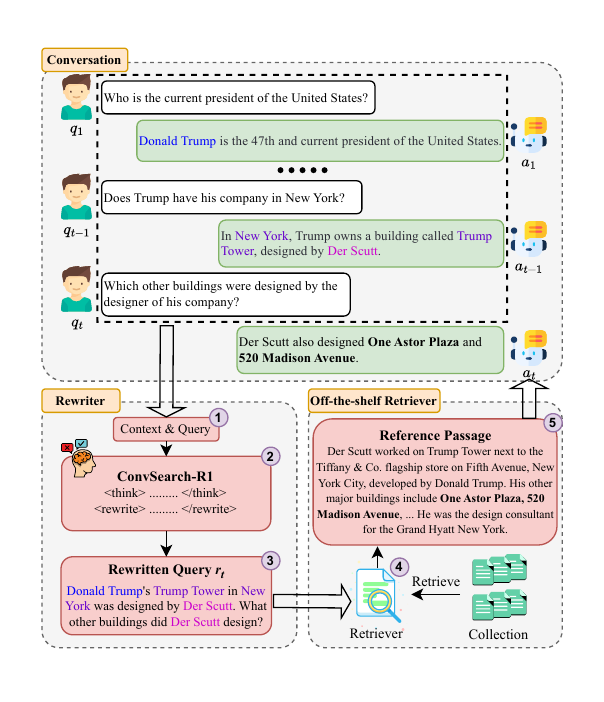}
  \caption{Illustration of the CQR task. Given a query and its context, the rewriter aims to reformulate the query into a stand-alone form, which facilitates the off-the-shelf retriever in finding the most relevant passage.}
  \vspace*{-15pt}
  \label{fig:cqr_demo}
\end{figure}

Conversational search aims to fulfill users' information needs through multi-turn interactions, unlike traditional information retrieval systems that only consider single-turn, keyword-based queries \citep{Joshi2017TriviaQAAL, kwiatkowski-etal-2019-natural}. In these multi-turn scenarios, conversational queries often contain ambiguity, omission, and coreference, making it difficult for existing retrieval methods to accurately capture user intent \citep{Anantha2020OpenDomainQA, Qu2020OpenRetrievalCQ, Gao2022NeuralAT}.
Given the challenges and computational costs of training multi-turn retrievers, Conversational Query Reformulation (CQR) \cite{elgohary-etal-2019-unpack,Vakulenko2020QuestionRF,Yu2020FewShotGC} has emerged as a practical solution, transforming context-dependent queries into self-contained forms that can be effectively processed by off-the-shelf retrievers (As shown in Figure \ref{fig:cqr_demo}). 

Existing CQR approaches have explored various strategies to address conversational search challenges. Some methods rely on explicit rewrites as supervision, obtained through either human annotations \cite{Lin2020ConversationalQR, Vakulenko2020QuestionRF, Tredici2021QRforODCQA,Vakulenko2021DBLP} or knowledge distillation \cite{mao-etal-2023-LLM4CS, mo-etal-2024-CHIQ} from large language models (LLMs) like ChatGPT \cite{achiam2023openai}.
More recent approaches attempt to leverage retrieval signals for preference optimization, though still require externally annotated data for initialization and generally limit exploration to pairwise preferences rather than optimizing for actual ranking improvements \cite{mo-etal-2023-convgqr, jang-etal-2024-itercqr, zhang-etal-2024-AdaQR, yoon-etal-2025-RetPO, lai-etal-2025-adacqr}.
These approaches suffer from two critical constraints: (1) the high dependency on costly and inconsistent external sources for high-quality annotation; and (2) insufficient alignment between the rewriting model and the downstream retriever. The fundamental challenge remains: how to enable query reformulation models to effectively align with retrievers without explicitly annotated reference rewrites, through self-exploration guided by retrieval feedback.

To address these challenges, we propose ConvSearch-R1, a novel self-driven framework that completely eliminates the dependency on external rewrite supervision. Leveraging retrieval ranking signals as rewards, the model self-discovers effective rewrites through iterative exploration and exploitation.
Specifically, we design a two-stage framework: (1) Self-Driven Policy Warm-Up (SDPWU), 
which addresses the cold-start problem by leveraging the model's few-shot reasoning capabilities combined with retrieval ranking signals to self-distill high-quality rewrite data without external supervision; and (2) 
Retrieval-Guided Reinforcement Learning, which further aligns the rewrite model with the retriever through Group Relative Policy Optimization (GRPO) \cite{Shao2024DeepSeekMathPT}. With a carefully designed rank-incentive reward shaping, ConvSearch-R1 addresses the sparsity issue in conventional retrieval metrics (like Recall@K and NDCG@K) \cite{Jiang2025DeepRetrievalHR}, providing smoother learning signals rather than binary or highly skewed outcomes, enabling stable and efficient exploration of the vast reformulation space.

We validate the effectiveness and generalizability of ConvSearch-R1 through extensive experiments using 3B parameter models. Compared to approaches employing 7B models \citep{zhang-etal-2024-AdaQR, yoon-etal-2025-RetPO}, our method is not only more cost-efficient but also delivers even better performance. ConvSearch-R1 achieves state-of-the-art performance on two widely-used conversational search datasets, TopiOCQA \citep{Adlakha2021TopiOCQAOC} and QReCC \citep{anantha-etal-2021-qrecc}. Notably, on the more challenging TopiOCQA dataset under dense retrieval, ConvSearch-R1, using Llama3.2-3B and Qwen2.5-3B as backbones, improves by 10.3\% and 10.7\% on average across all metrics, respectively, compared to previous state-of-the-art results. This demonstrates that, even without human rewrites or external distilled data, relying solely on self-distillation combined with reinforcement learning (RL) under the reasoning mode enables the model to perform effectively on the CQR task.

Our contributions are summarized as follows: 
\begin{itemize}

\item We propose ConvSearch-R1, the first conversational query rewriting approach that completely eliminates dependency on external rewrite supervision, enabling effective alignment with off-the-shelf retrievers without costly human annotations.
\item We introduce a novel two-stage alignment framework comprising self-driven policy warm-up and rank-incentive reward shaping that effectively addresses the cold-start problem and reward sparsity challenges inherent in retrieval-aligned optimization.
\item Our extensive experiments across multiple datasets and retrievers demonstrate substantial performance improvements over state-of-the-art methods, particularly on the challenging TopiOCQA dataset, where ConvSearch-R1 achieves over 10\% average improvement across all metrics while using smaller language models and no external supervision.

\item To facilitate future research in this area, we make datasets, code, and models available at  \url{https://github.com/BeastyZ/ConvSearch-R1}.
\end{itemize}

\section{Related Work}
\subsection{Conversational Search}

Conversational search \cite{Gao2022NeuralAT} is a way of searching for information by having a natural, back-and-forth dialogue with a search system, similar to talking with a person. The core challenge of conversational search lies in addressing omission, ambiguity, and coreference present in the current query \citep{Anantha2020OpenDomainQA, Qu2020OpenRetrievalCQ, Gao2022NeuralAT}. Existing approaches to conversational search can be broadly categorized into two main types: conversational dense retrieval (CDR) and CQR. 
For CDR, many existing methods \citep{mao-etal-2024-chatretriever, kim-kim-2022-saving, Mo2024HistoryAwareCD} use a substantial amount of annotated session-passage pairs to fine-tune an ad-hoc retriever into a context-aware conversational retriever, a process that is costly and may not fully take  the advantages of off-the-shelf retrievers.
On the other hand, CQR leverages the strengths of existing ad-hoc retrieval systems by striving to transform context-dependent queries into stand-alone forms. Early studies \citep{Voskarides2020QueryRF, Lin2020ConversationalQR} primarily relied on human rewrites to endow models with query rewriting capability. With the advent of LLMs, some recent works have begun to utilize the power of LLMs for query rewriting. \citet{mao-etal-2023-LLM4CS} and \citet{Ye2023EnhancingCS} employ ChatGPT to perform query rewriting via a training-free, purely prompt-based method. Meanwhile, many approaches \citep{mo-etal-2023-convgqr, zhang-etal-2024-AdaQR, jang-etal-2024-itercqr, mo-etal-2024-CHIQ, lai-etal-2025-adacqr, yoon-etal-2025-RetPO} distill high-quality rewrites from ChatGPT or Llama to train rewriters that possess strong rewriting ability from the outset. 
In contrast, ConvSearch-R1 improves query rewriting capability through self-distillation and RL-based trial and error. ConvSearch-R1 not only significantly reduces the cost of obtaining rewrites but also achieves state-of-the-art performance.

\subsection{RLVR-based Retrieval}

Reinforcement learning from verifiable reward (RLVR) has recently emerged as a powerful approach for enhancing language models' capabilities without explicit supervision \citep{DeepSeekAI2025DeepSeekR1IR}. While recent work has integrated RLVR with retrieval mechanisms \citep{Jin2025SearchR1TL, Song2025R1SearcherIT, Chen2025ReSearchLT, sun2025zerosearch}, these efforts primarily focus on improving single-turn question answering by teaching LLMs to better utilize retrieval tools. In contrast, our approach leverages reasoning to optimize query reformulation specifically for conversational search contexts.
In the retrieval domain, \citet{Jiang2025DeepRetrievalHR} applies RLVR for query generation, but differs from our work in two key aspects: (1) it uses retrieval metrics directly as rewards, while we develop a specialized rank-incentive reward function, which significantly enhances retrieval performance; and (2) it addresses only single-turn scenarios, not the conversational challenges of omission, ambiguity, and coreference resolution.
Regarding conversational search specifically, existing RL approaches \citet{Wu2021CONQRRCQ, chen-etal-2022-reinforced} rely on human rewrites for reward computation, whereas our method eliminates external supervision requirements while incorporating reasoning capabilities into the rewriter.

\section{ConvSearch-R1}
\begin{figure*}[t]
    \centering
    \includegraphics[width=\linewidth]{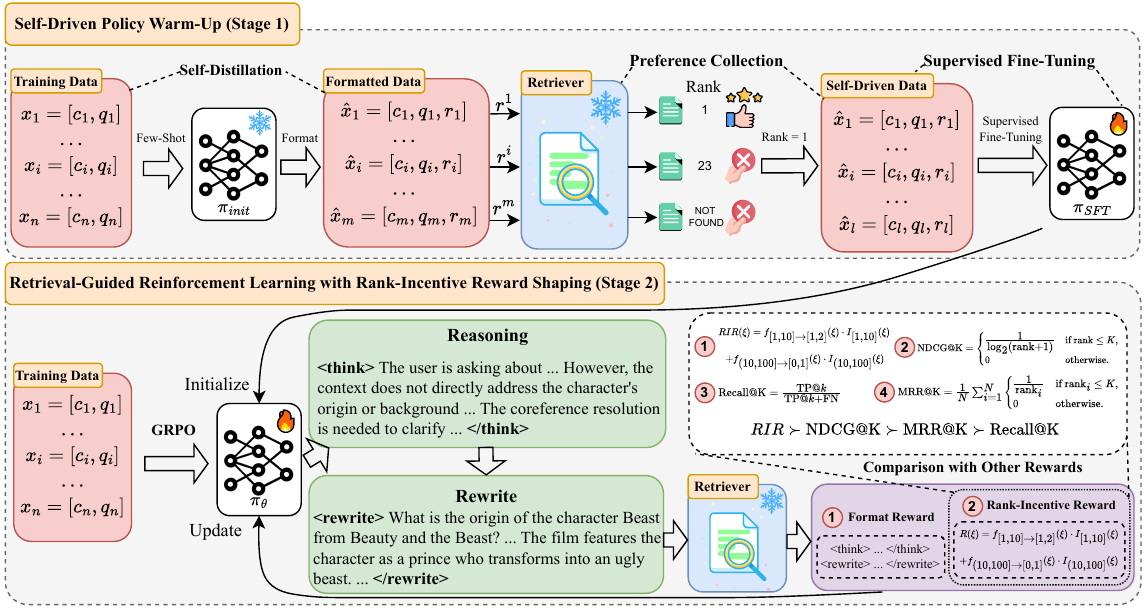}
    \caption{Overview of ConvSearch-R1. In Stage 1, we self-distilled a set of high-quality data using few-shot learning and obtained the corresponding SFT model. In Stage 2, we further improved the rewriter's performance via RL by refining the reward function.}
    \label{fig:framework}
\end{figure*}

\subsection{Task Formulation}
A conversational search session is defined as a sequence of alternating user queries and system answers:
\(S = \{(q_1, a_1), (q_2, a_2), \ldots, (q_t, a_t)\}\),
where \(q_t\) denotes the user query at turn \textit{t}, and \(a_t\) denotes the corresponding system answer. At turn \textit{t}, the user issues a query \(q_t\), which is potentially dependent on the previous conversational history 
\(C_{t-1} = \{(q_1, a_1), (q_2, a_2), \ldots, (q_{t-1}, a_{t-1})\}\).
The task of CQR is, given the current query \(q_t\) and the preceding conversational history \(C_{t-1}\), to generate a context-independent reformulated query 
\(r_t = f(q_t, C_{t-1})\), 
where \(f(\cdot)\) is the reformulation function that leverages both previous queries and system answers to resolve omission, ambiguity, and coreference in \(q_t\).
The optimal reformulation function \(f^*\) can be defined as:
\[f^* = \arg\max_{f} \mathbb{E}_S[\delta(\mathcal{T}(r_t), \hat{p}_t)],\]
where \(\hat{p}_t\) is the gold-standard passage for \(q_t\) in the collection, \(\mathcal{T}(\cdot)\) is the retriever that returns passages given a query, and \(\delta(\cdot,\cdot)\) is a matching function that measures the relevance between the retrieved passage and the gold passage.

\subsection{Overview}
ConvSearch-R1 employs a reasoning-based approach to enable the rewriter to fully grasp the omission, ambiguity, and coreference present in the current user query, thereby generating rewrites that are both context-independent and semantically rich. As illustrated in Figure \ref{fig:framework}, ConvSearch-R1 does not require any external supervised data (reference rewrite) and consists of the following stages: (1) Self-distill a set of format-compliant data from the \(\pi_{init}\) \footnote{\(\pi_{init}\) refers to Qwen2.5-3B or Llama3.2-3B used in the paper.} through few-shot learning, aligning it with the preferences of the retriever. Only the top-ranked (rank-1) rewrites are retained and used to fine-tune the \(\pi_{init}\), thus getting the \(\pi_{SFT}\) with initial ability to follow the desired output format and perform query rewriting; (2) Further improve the \(\pi_{SFT}\) using RL with rank-incentive reward shaping, yielding the final rewriter \(\pi_{\theta}\). 
The output format of the rewriter is required to strictly adhere to the following structure: \textit{<think> reasoning process here </think>\textbackslash n<rewrite> rewrite here </rewrite>}.

\subsection{Self-Driven Policy Warm-Up}
\label{sec:SDPWU}

Following \citet{Chu2025SFTMR}'s findings on the crucial role of supervised fine-tuning (SFT) in stabilizing output formats for effective RL training, we introduce a self-driven policy warm-up strategy that avoids costly knowledge distillation from more powerful LLMs used in previous approaches. 

\paragraph{Self-Distillation.} 
We begin by generating rewrites using few-shot prompting applied to our initial model $\pi_{init}$: $D^{d} = \{ y_i = \pi_{init}(x_i, \rho, instruction) \mid x_i \in D \}$, where $D$ is the original dataset, $x_i$ is the $i$-th sample in the $D$, $\rho$ is a fixed set of few-shot examples, and $D^{d}$ is the distilled dataset.
We then filter these outputs to obtain format-compliant samples: $D^{f} = \{ y \in D^{d} \mid g(y) = 1 \}$, where \(g:Y\to\{0,1\}\) is a format validation function indicating whether the output meets the required constraints.

\paragraph{Preference Collection and Supervised Fine-Tuning.} We identify high-quality rewrites by retaining only samples whose rewrites rank the gold passage at position 1, creating our self-driven data (SD-DATA). Each sample in SD-DATA is represented as a triplet [$c$,$q$,$r$], containing conversational history \textit{c}, current query \textit{q}, and the reasoning-rewrite pair \textit{r}. For more details about data collection, see Appendix \ref{appendix:data-collection}. Finally, we fine-tune $\pi_{init}$ on SD-DATA to maximize the likelihood of ground-truth outputs, producing $\pi_{SFT}$ with fundamental capabilities in format adherence, reasoning, and query rewriting. Through this process, our model learns to first generate an appropriate reasoning process and then produce a context-independent rewrite, conditioned on both the conversational history and the generated reasoning.

\subsection{Retrieval-Guided Reinforcement Learning}
\label{sec:rirs}

\paragraph{Rank-Incentive Reward Shaping.} The design of reward functions is critical in RL, directly impacting the learning effectiveness of policy models \citep{Rati2021reward_desing}. Unlike \citet{Jiang2025DeepRetrievalHR} which directly uses retrieval metrics as reward signals, we propose Rank-Incentive Reward Shaping (RIRS) to address the reward sparsity problem. As shown in Figure \ref{fig:reward_sparsity} in the Appendix, directly using metrics like MRR@3 and NDCG@3 as rewards leads to severe reward sparsity, hindering effective model learning.

RIRS utilizes retrieval ranking positions to create a more informative reward signal rather than relying on binary retrieval metrics. 
Considering users typically pay more attention to top positions, RIRS implements a piecewise reward function that allocates differentiated reward intensities—assigning higher rewards for top positions (1-10) while maintaining proportionally smaller but meaningful rewards for moderate positions (11-100).
This approach preserves semantic thresholds of retrieval quality while ensuring dense feedback signals throughout the policy optimization process. 
The RIRS reward function is formally defined as:

\vspace{-1.5em}
\begin{equation}
\begin{split}
    R(\xi) = f_{[1,10] \rightarrow [1,2]}(\xi ) \cdot I_{[1,10]}(\xi) \\+ f_{(10,100] \rightarrow [0,1]}(\xi) \cdot I_{(10,100]}(\xi),
\end{split}
\end{equation}
where $f_{A \rightarrow B}$ represents a function mapping from interval \textit{A} to interval \textit{B}, $I_A(\xi )$ is the indicator function, which equals 1 when $\xi$ is in set \textit{A} and 0 otherwise, $\xi $ is the rank variable.

Considering format correctness, the complete reward function is:

\vspace{-1.5em}
\begin{equation}
R(\xi,\phi)=R(\xi)\cdot I(\phi=1)+\delta\cdot I(\phi=0),
\end{equation}
where $\phi \in \{0,1\}$ is the format compliance indicator, and $\delta =- 0.1$ is the penalty term for format non-compliance.

\paragraph{GRPO.} Equipped with the rewriter \(\pi_{SFT}\) obtained from Stage 1 SDPWU and the carefully designed reward function (Rank-Incentive Reward), we adopt GRPO \citep{Shao2024DeepSeekMathPT} as the specific RL algorithm. GRPO is an efficient algorithm that eliminates the need for an explicit reward model and value model. 
Through RL with RIRS, an existing \(\pi_{\theta}\) is incentivized to explore the solution space while rollouts \(\{r_i\}_{i=1} \sim \pi_{\theta_{old}}(\cdot|x)\) are used to maximize retrieval performance.
The optimization objective is formulated as:

\vspace{-1em}
\begin{small}
\begin{equation}
\begin{aligned}
J_{GRPO}(\theta) = &\mathbb{E}\Big[
\sum_{i} \min\Big(\frac{\pi_{\theta}(r_i|x)}{\pi_{\theta_{old}}(r_i|x)}A_i, \
\text{clip}\Big(\frac{\pi_{\theta}(r_i|x)}{\pi_{\theta_{old}}(r_i|x)}, 
\\& 1-\epsilon, 1+\epsilon\Big)A_i\Big)
- \beta D_{KL}(\pi_{\theta}||\pi_{SFT})\Big],
\end{aligned}
\end{equation}
\end{small}
where 
\begin{small}
$ A_i = \frac{R(\xi_i,\phi_i) - \text{mean}({R(\xi,\phi)})}{\text{std}({R(\xi,\phi)})} $
\end{small}
represents the normalized advantage of the $i$-th rewritten query within the current group, calculated using our RIRS reward function. The parameter $\epsilon$ controls the clipping threshold, while $\beta$ regulates the KL divergence penalty.

\section{Experimental Setup}
\noindent \textbf{Settings.}\quad We follow prior work \citep{yoon-etal-2025-RetPO} in configuring the datasets, retrievers, and metrics. For datasets, we employ TopiOCQA \citep{Adlakha2021TopiOCQAOC} and QReCC \citep{anantha-etal-2021-qrecc}, which are widely used in the conversational search task. For retrievers, we utilize BM25 as the sparse retriever for all experiments and ANCE \citep{Xiong2020ANCE} as the dense retriever for all experiments, where ANCE is trained on the MS-MARCO \citep{Campos2016MSMARCO} passage retrieval tasks. Notably, we do not train any retrievers in our experiments. For metrics, we adopt MRR@3, NDCG@3, and Recall@K (referred to as R@K in this paper) for evaluation. See Appendix \ref{appendix:data_details} \& \ref{appendix:metric_details} for more details.

\smallskip

\noindent \textbf{Baselines
.}\quad We consider four categories of baselines in our experiments. The first category comprises basic baselines that do not involve any query rewriting optimization, including Human Rewrite, Raw, DS-R1-Distill-Qwen-7B, Llama3.2-3B, and Qwen2.5-3B. The second category consists of baselines that fine-tune small-scale models (e.g., T5 and BERT), including QuReTec~\citep{Voskarides2020QueryRF}, T5QR~\citep{Lin2020ConversationalQR}, CONQRR~\citep{Wu2021CONQRRCQ}, ConvGQR~\citep{mo-etal-2023-convgqr}, EDIRCS~\citep{mao-etal-2023-search}, IterCQR~\citep{jang-etal-2024-itercqr}, ADACQR~\citep{lai-etal-2025-adacqr}, and CHIQ-Fusion~\citep{mo-etal-2024-CHIQ}. The third category involves baselines that fine-tune LLMs (e.g., Llama2-7B) , including RETPO~\citep{yoon-etal-2025-RetPO} and AdaQR~\citep{zhang-etal-2024-AdaQR}. The fourth category is training-free, leveraging prompt-based methods with ChatGPT for query rewriting. This category includes LLM4CS~\citep{mao-etal-2023-LLM4CS} and InfoCQR~\citep{Ye2023EnhancingCS}. A detailed description of each aforementioned baseline is presented in Appendix \ref{appendix:baselines_details}.
 
\smallskip

\noindent \textbf{Implementation Details.}\quad We employ Llama3.2-3B and Qwen2.5-3B as the backbone models for our rewrite. 
For training, we utilize verl \citep{Sheng2024verl}, a flexible and efficient RLHF framework. The BM25 retriever is implemented using Pyserini \citep{Lin2021Pyserini}, while the ANCE retriever is built with Faiss \citep{Johnson2017faiss}. Evaluation metrics are computed with pytrec\_eval \citep{Gysel2018Pytrec_eval}. More implementation details can be found in Appendix \ref{appendix:imple_details}.

\section{Results and Analysis}
\subsection{Main Results}
\label{sec:main_results}

\definecolor{mygreen}{HTML}{d9eddf}
\definecolor{mygrey}{HTML}{EBEBEB}

\newcommand{\pub}[1]{{\color{gray}{\small\textit{({#1})\!}}}}

\begin{table*}[!th]
    \resizebox{\textwidth}{!}{
    \renewcommand\tabcolsep{4pt}
        \begin{tabular}{clccccccccccccc}\toprule
         \multirow{2}{*}{\textbf{Type}} &\multirow{2}{*}{\textbf{Method}} &\multirow{2}{*}{\textbf{NE}} & \multirow{2}{*}{\textbf{NH}} &\multicolumn{4}{c}{\textbf{TopiOCQA}} &\multicolumn{4}{c}{\textbf{QReCC}} &\multirow{2}{*}{\textbf{Avg}} \\\cmidrule{5-12}
        & & & &\textbf{MRR@3} &\textbf{NDCG@3} &\textbf{R@10} &\textbf{R@100} &\textbf{MRR@3} &\textbf{NDCG@3} &\textbf{R@10} &\textbf{R@100} & \\\midrule
        \multirow{21}{*}{{\rotatebox{90}{\textbf{Sparse (BM25)}}}} &Human Rewrite &- &- &- &- &- &- &39.8 &36.3 &62.7 &\textbf{98.5} &- \\
        &Raw &- &- &2.1 &1.8 &4.0 &9.2 &6.5 &5.5 &11.1 &21.5 &7.7 \\
        &DS-R1-Distill-Qwen-7B &- &- &10.0 &8.6 &18.8 &38.3 &29.4 &27.3 &44.0 &63.9 &30.0 \\\cmidrule(lr){2-13}
        &QuReTeC \pub{SIGIR 2020} &\cmark &\xmark &8.5 &7.3 &16.0 &- &34.0 &30.5 &55.5 &- &- \\
        &T5QR &\cmark &\xmark &11.3 &9.8 &22.1 &44.7 &33.4 &30.2 &53.8 &86.1 &36.4 \\
        &CONQRR \pub{EMNLP 2022} &\xmark &\xmark &- &- &- &- &38.3 &- &60.1 &88.9 &- \\
        &ConvGQR \pub{ACL 2023} &\cmark &\xmark &12.4 &10.7 &23.8 &45.6 &44.1 &41.0 &64.4 &88.0 &41.3 \\
        &EDIRCS \pub{ACL 2023} &\xmark &\xmark &- &- &- &- &41.2 &- &62.7 &90.2 &- \\
        &IterCQR \pub{NAACL 2024} &\xmark &\cmark &16.5 &14.9 &29.3 &54.1 &46.7 &44.1 &64.4 &85.5 &44.4 \\
        &ADACQR \pub{COLING 2025} &\xmark &\cmark &28.3 &26.5 &48.9 &71.2 &55.1 &52.5 &76.5 &93.7 &56.6 \\
        &CHIQ-Fusion \pub{EMNLP 2024} &\xmark &\cmark &25.6 &23.5 &44.7 &- &54.3 &51.9 &\textbf{78.5} &- &- \\
        &RETPO \pub{NAACL 2025} &\xmark &\cmark &28.3 &26.5 &48.3 &73.1 &50.0 &47.3 &69.5 &89.5 &54.1 \\
        &AdaQR \pub{EMNLP 2024} &\xmark &\cmark &20.3 &18.0 &37.1 &66.2 &50.6 &48.0 &69.6 &- &- \\
        &LLM4CS \pub{EMNLP 2023} &\xmark &\cmark &18.9 &17.7 &33.7 &- &47.8 &45.0 &69.1 &- &- \\
        &InfoCQR \pub{EMNLP 2023} &\xmark &\cmark &- &- &- &- &48.9 &46.3 &66.4 &- &- \\
        \cmidrule(lr){2-13}
        &LLama3.2-3B &- &- &4.8 &4.0 &8.6 &20.7 &22.5 &21.0 &33.7 &49.8 &20.6 \\
        &\cellcolor{mygreen} + ConvSearch-R1(ours) &\cellcolor{mygreen}\cmark &\cellcolor{mygreen}\cmark &\cellcolor{mygreen}\textbf{37.8} &\cellcolor{mygreen}\textbf{36.2} &\cellcolor{mygreen}\textbf{59.6} &\cellcolor{mygreen}\textbf{80.1} &\cellcolor{mygreen}\textbf{55.9} &\cellcolor{mygreen}\textbf{54.3} &\cellcolor{mygreen}77.2 &\cellcolor{mygreen}89.0 &\cellcolor{mygreen}\textbf{61.3} \\
        & & & &\cellcolor{mygrey} \textit{+9.5}&\cellcolor{mygrey}\textit{+9.7} &\cellcolor{mygrey}\textit{+10.7} &\cellcolor{mygrey}\textit{+7.0} &\cellcolor{mygrey}\textit{+0.8} &\cellcolor{mygrey}\textit{+1.8} &\cellcolor{mygrey}\textit{-1.3} &\cellcolor{mygrey}\textit{-4.7} &\cellcolor{mygrey}\textit{+4.7} \\
        \cmidrule(lr){2-13}
        &Qwen2.5-3B &- &- &8.8 &7.5 &17.3 &36.1 &27.3 &25.0 &42.3 &64.6 &28.6 \\
        &\cellcolor{mygreen} + ConvSearch-R1(ours) &\cellcolor{mygreen}\cmark &\cellcolor{mygreen}\cmark &\cellcolor{mygreen}\textbf{35.2} &\cellcolor{mygreen}\textbf{33.5} &\cellcolor{mygreen}\textbf{57.8} &\cellcolor{mygreen}\textbf{79.9} &\cellcolor{mygreen}\textbf{56.5} &\cellcolor{mygreen}\textbf{54.8} &\cellcolor{mygreen}76.3 &\cellcolor{mygreen}88.1 &\cellcolor{mygreen}\textbf{60.3} \\
        & & & &\cellcolor{mygrey}\textit{+6.9} &\cellcolor{mygrey}\textit{+7.0} &\cellcolor{mygrey}\textit{+8.9} &\cellcolor{mygrey}\textit{+6.8} &\cellcolor{mygrey}\textit{+1.4} &\cellcolor{mygrey}\textit{+2.3} &\cellcolor{mygrey}\textit{-2.2} &\cellcolor{mygrey}\textit{-5.6} &\cellcolor{mygrey}\textit{+3.7} \\
        \midrule
        \multirow{21}{*}{{\rotatebox{90}{\textbf{Dense (ANCE)}}}} &Human Rewrite &- &- &- &- &- &- &38.4 &35.6 &58.6 &78.1 &- \\
        &Raw &- &- &4.1 &3.8 &7.5 &13.8 &10.2 &9.3 &15.7 &22.7 &10.9 \\
        &DS-R1-Distill-Qwen-7B &- &- &21.0 &19.9 &36.4 &53.4 &28.4 &25.9 &43.9 &59.0 &36.0 \\\cmidrule(lr){2-13}
        &QuReTeC \pub{SIGIR 2020} &\cmark &\xmark &11.2 &10.5 &20.2 &- &35.0 &32.6 &55.0 &- &- \\
        &T5QR &\cmark &\xmark &23.0 &22.2 &37.6 &54.4 &34.5 &31.8 &53.1 &72.8 &41.2 \\
        &CONQRR \pub{EMNLP 2022} &\xmark &\xmark &- &- &- &- &41.8 &- &65.1 &84.7 &- \\
        &ConvGQR \pub{ACL 2023} &\cmark &\xmark &25.6 &24.3 &41.8 &58.8 &42.0 &39.1 &63.5 &81.8 &47.1 \\
        &EDIRCS \pub{ACL 2023} &\xmark &\xmark &- &- &- &- &42.1 &- &65.6 &\textbf{85.3} &- \\
        &IterCQR \pub{NAACL 2024} &\xmark &\cmark &26.3 &25.1 &42.6 &62.0 &42.9 &40.2 &65.5 &84.1 &48.6 \\
        &ADACQR \pub{COLING 2025} &\xmark &\cmark &38.5 &37.6 &58.4 &75.0 &45.8 &42.9 &67.3 &83.8 &56.2 \\
        &CHIQ-Fusion \pub{EMNLP 2024} &\xmark &\cmark &38.0 &37.0 &61.6 &- &47.2 &44.2 &\textbf{70.7} &- &- \\
        &RETPO \pub{NAACL 2025} &\xmark &\cmark &30.0 &28.9 &49.6 &68.7 &44 &41.1 &66.7 &84.6 &51.7 \\
        &AdaQR \pub{EMNLP 2024} &\xmark &\cmark &38.1 &36.6 &61.3 &79.9 &43.4 &40.8 &65.6 &- &- \\
        &LLM4CS \pub{EMNLP 2023} &\xmark &\cmark &27.7 &26.7 &43.3 &- &44.8 &42.1 &66.4 &- &- \\
        &InfoCQR \pub{EMNLP 2023} &\xmark &\cmark &- &- &- &- &43.9 &41.3 &65.6 &- &- \\\cmidrule(lr){2-13}
        &LLama3.2-3B &- &- &12.8 &11.7 &22.6 &38.1 &19.1 &17.4 &30.4 &44.1 &24.5 \\
        &\cellcolor{mygreen}+ ConvSearch-R1(ours) &\cellcolor{mygreen}\cmark &\cellcolor{mygreen}\cmark &\cellcolor{mygreen}\textbf{50.5} &\cellcolor{mygreen}\textbf{50.1} &\cellcolor{mygreen}\textbf{72.0} &\cellcolor{mygreen}\textbf{86.3} &\cellcolor{mygreen}\textbf{50.2} &\cellcolor{mygreen}\textbf{48.1} &\cellcolor{mygreen}70.6 &\cellcolor{mygreen}82.8 &\cellcolor{mygreen}\textbf{63.8} \\
        & & & &\cellcolor{mygrey}\textit{+12.0} &\cellcolor{mygrey}\textit{+12.5} &\cellcolor{mygrey}\textit{+10.4} &\cellcolor{mygrey}\textit{+6.4} &\cellcolor{mygrey}\textit{+3.0} &\cellcolor{mygrey}\textit{+3.9} &\cellcolor{mygrey}\textit{-0.1} &\cellcolor{mygrey}\textit{-2.5} &\cellcolor{mygrey}\textit{+7.6} \\
        \cmidrule(lr){2-13}
        &Qwen2.5-3B &- &- &19.2 &18.3 &33.0 &46.7 &29.3 &27.2 &44.2 &59.0 &34.6 \\
        &\cellcolor{mygreen}+  ConvSearch-R1(ours) &\cellcolor{mygreen}\cmark &\cellcolor{mygreen}\cmark &\cellcolor{mygreen}\textbf{51.4} &\cellcolor{mygreen}\textbf{51.3} &\cellcolor{mygreen}\textbf{72.0} &\cellcolor{mygreen}\textbf{85.7} &\cellcolor{mygreen}\textbf{49.7} &\cellcolor{mygreen}\textbf{47.7} &\cellcolor{mygreen}69.8 &\cellcolor{mygreen}81.6 &\cellcolor{mygreen}\textbf{63.7} \\
        & & & &\cellcolor{mygrey}\textit{+12.9} &\cellcolor{mygrey}\textit{+13.7} &\cellcolor{mygrey}\textit{+10.4} &\cellcolor{mygrey}\textit{+5.8} &\cellcolor{mygrey}\textit{+2.5} &\cellcolor{mygrey}\textit{+3.5} &\cellcolor{mygrey}\textit{-0.9} &\cellcolor{mygrey}\textit{-3.7} &\cellcolor{mygrey}\textit{+7.5} \\
        \bottomrule
        \end{tabular}
    }
    \caption{Results of both dense and sparse retrieval on TopiOCQA and QReCC. \textbf{NE} denotes no external distillation, indicating that no external data was distilled from open-source or closed-source LLMs. \textbf{NH} stands for no human, meaning that no human rewrites were utilized. \textbf{Bold} indicates the best results, and the rest of the tables follow the same convention. \colorbox{mygrey}{Grey} indicates the improvements over SOTA baselines.}
    \label{tab:main_res}
\end{table*}

Table \ref{tab:main_res} shows the retrieval performance of various methods in dense and sparse settings, leading to the following conclusions:

\begin{table*}[!th]
    \resizebox{\textwidth}{!}{
        \begin{tabular}{clccccccccc}\toprule
        \multirow{2}{*}{\textbf{Type}} &\multirow{2}{*}{\textbf{Method}} &\multicolumn{4}{c}{\textbf{TopiOCQA}} &\multicolumn{4}{c}{\textbf{QReCC}} \\\cmidrule{3-10}
        & &\textbf{MRR@3} &\textbf{NDCG@3} &\textbf{R@10} &\textbf{R@100} &\textbf{MRR@3} &\textbf{NDCG@3} &\textbf{R@10} &\textbf{R@100} \\\midrule
        \multirow{5}{*}{\textbf{Sparse(BM25)}} &IterCQR &13.7 &12.2 &- &- &44.9 &42.4 &- &- \\
        &ADACQR &14.0 &12.6 &- &- &- &- &- &- \\
        &RETPO &17.2 &- &32.0 &59.1 &40.1 &- &62.2 &86.5 \\
        \cmidrule(lr){2-10}
        &ConvSearch-R1(ours, Llama3.2-3B) &\textbf{27.6} &\textbf{25.8} &\textbf{48.3} &\textbf{74.5} &\textbf{49.9} &\textbf{47.6} &\textbf{72.0} &\textbf{87.2} \\
        &ConvSearch-R1(ours, Qwen2.5-3B) &\textbf{24.0} &\textbf{22.1} &\textbf{42.4} &\textbf{69.2} &\textbf{46.2} &\textbf{43.9} &\textbf{68.5} &83.8 \\
        \midrule
        \multirow{5}{*}{\textbf{Dense(ANCE)}} &IterCQR &17.8 &16.4 &- &- &40.1 &37.4 &- &- \\
        &ADACQR &20.1 &18.6 &- &- &- &- &- &- \\
        &RETPO &23.2 &- &40.0 &59.4 &40.9 &- &61.9 &\textbf{79.9} \\
        \cmidrule(lr){2-10}
        &ConvSearch-R1(ours, Llama3.2-3B) &\textbf{36.8} &\textbf{35.4} &\textbf{58.8} &\textbf{78.4} &\textbf{42.8} &\textbf{40.2} &\textbf{63.5} &78.7 \\
        &ConvSearch-R1(ours, Qwen2.5-3B) &\textbf{35.2} &\textbf{34.2} &\textbf{57.0} &\textbf{75.4} &\textbf{41.5} &\textbf{39.0} &61.8 &76.6 \\
        \bottomrule
        \end{tabular}
    }
    \caption{Performance on unseen datasets.}
    \label{tab:generalization}
\end{table*}

\textbf{Employing a two-stage alignment framework further enhances the final performance in the absence of external supervised data.} Directly using the current user query (\texttt{Raw}) for retrieval yields the poorest performance, underscoring the necessity of CQR. Notably, \texttt{Human Rewrite} not only fails to deliver optimal results but also performs significantly worse than many other baselines. This suggests that, in addition to the high annotation costs, human rewrites are not aligned with retriever preferences.
Furthermore, directly applying the reasoning model (i.e., \texttt{DS-R1-Distill-Qwen-7B}) to conversational search yields even worse results than \texttt{Human Rewrite}. This indicates that straightforward general-domain reasoning is not a silver bullet, and even models equipped with long chains-of-thought capability (i.e., R1-like models) struggle to excel in conversational search scenarios.
Methods that rely solely on human rewrites as supervision signals (e.g., \texttt{QuReTeC}, \texttt{T5QR}) exhibit the lowest performance among all baselines, highlighting the clear limitations of human rewrites. In contrast, approaches that utilize retrieval signals for supervision (e.g., \texttt{CONQRR}, \texttt{EDIRCS}, \texttt{IterCQR}, \texttt{ADACQR}, \texttt{RETPO}, \texttt{AdaQR}) achieve notable performance improvements, yet still fall short of our method. These results collectively demonstrate the effectiveness of our proposed two-stage alignment framework. Notably, compared to the baselines, ConvSearch-R1 achieves new state-of-the-art performance across most experimental settings without the need for any external supervised data.

\textbf{While the baselines leverage larger language models to achieve competitive performance, our 3B rewriter demonstrates superior results.} Many studies (e.g., \texttt{RETPO}, \texttt{AdaQR}, \texttt{LLM4CS}, \texttt{InfoCQR}) have attempted to tackle the CQR task by leveraging larger language models (such as ChatGPT). However, our 3B-parameter model significantly outperforms these approaches and achieves new state-of-the-art results. Both \texttt{RETPO} and \texttt{AdaQR} employ Direct Preference Optimization \citep{Rafailov2023DirectPO} to align the retriever’s preferences, but this strategy overlooks the potential of enabling the model to fit preferences through trial and error. In contrast, ConvSearch-R1 utilizes RL with a Rank-Incentive Reward, which allows the model to explore and refine preference alignment through iterative trial and error, ultimately leading to a more optimal solution for retriever preference modeling with reduced parameter usage.

\begin{figure}[!t]
    \centering
    \begin{subfigure}{0.45\textwidth}
        \centering
        \includegraphics[width=\linewidth]{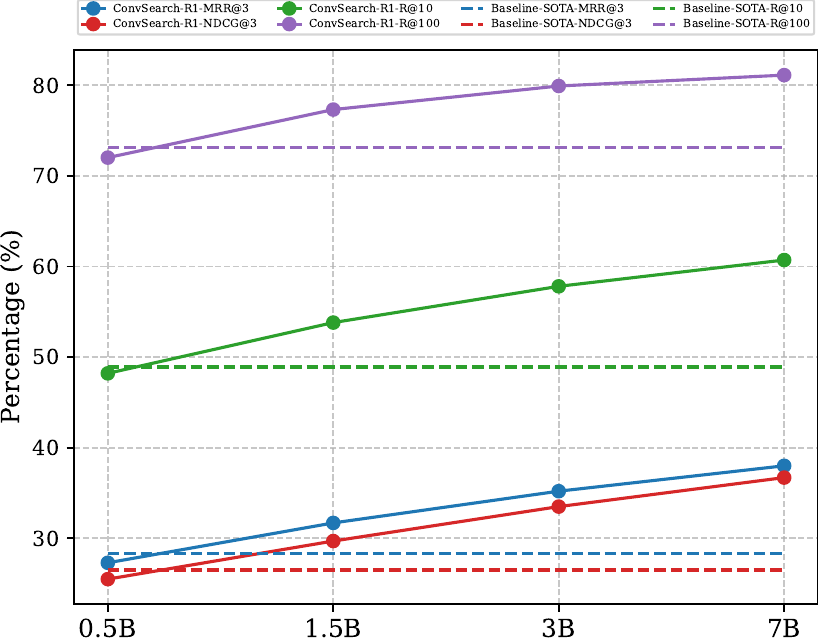}
        \caption{Comparison on BM25}
    \end{subfigure}
    \hfill
    \begin{subfigure}{0.43\textwidth}
        \centering
        \includegraphics[width=\linewidth]{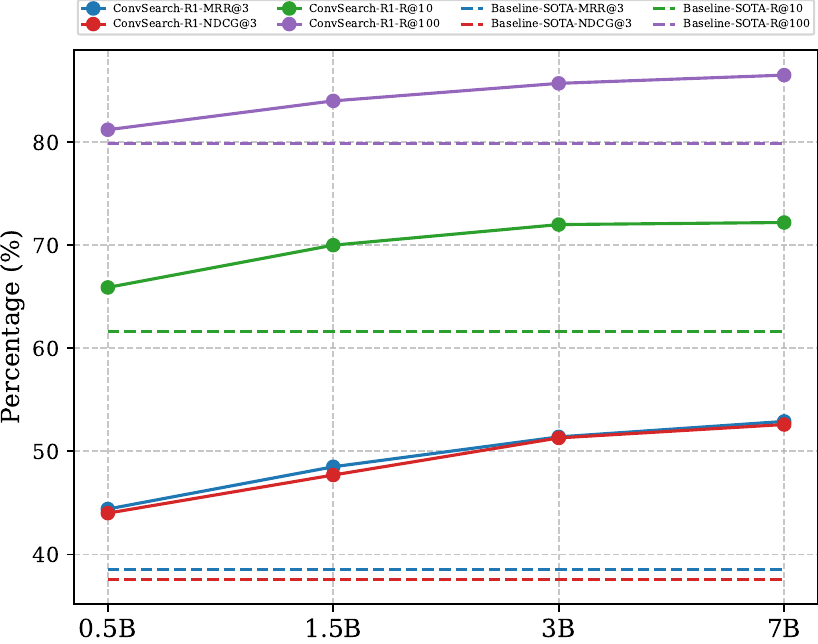}
        \caption{Comparison on ANCE}
    \end{subfigure}
    
    \caption{Model scale analysis on TopiOCQA.}
    \label{fig:model_scale}
    \vspace{-1em}
\end{figure}

\begin{figure}[t]
    \centering
    \begin{subfigure}{0.43\textwidth} 
        \centering
        \includegraphics[width=\linewidth]{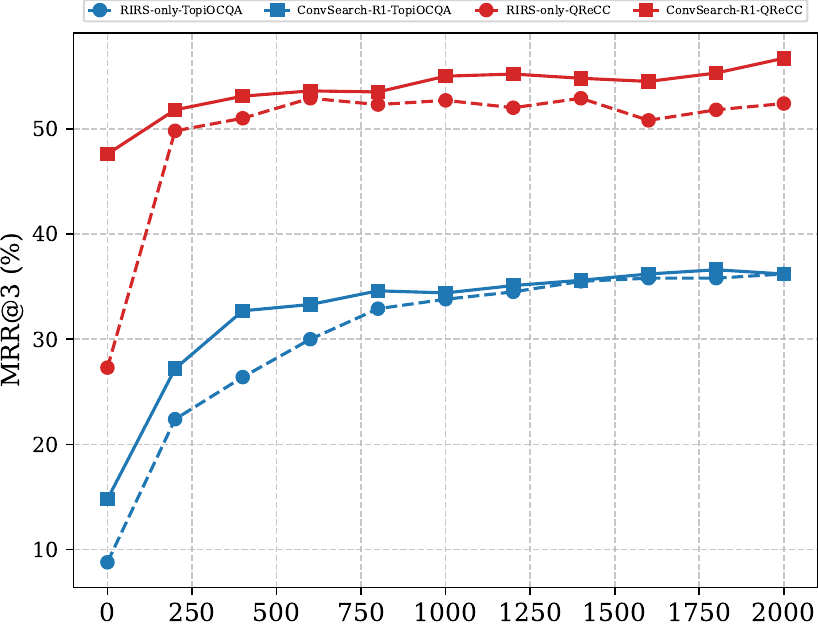}
        \caption{Comparison on BM25}
    \end{subfigure}
    \hfill
    \begin{subfigure}{0.43\textwidth}
        \centering
        \includegraphics[width=\linewidth]{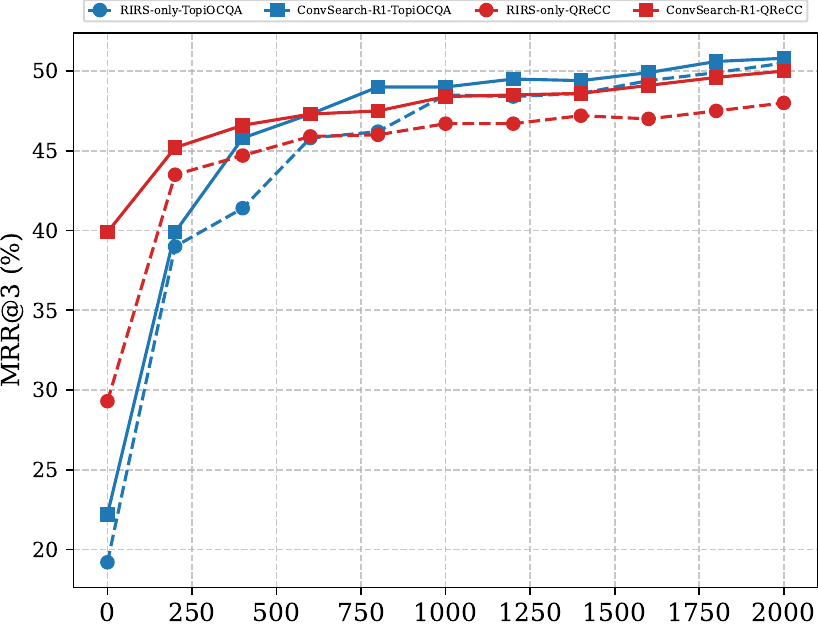}
        \caption{Comparison on ANCE}
    \end{subfigure}
    
    \caption{Warm-up analysiss using Qwen2.5-3B.}
    \label{fig:warmup_ana}
    \vspace{-1em}
\end{figure}

\subsection{Generalization on Unseen Datasets}
\label{sec:generalizaiton}

\begin{table*}[!th]
    \resizebox{\textwidth}{!}{
        \begin{tabular}{clccccccccccc}\toprule
        \multirow{2}{*}{\textbf{Type}} &\multirow{2}{*}{\textbf{Method}} &\multirow{2}{*}{\textbf{SDPWU}} &\multirow{2}{*}{\textbf{RIRS}} &\multicolumn{4}{c}{\textbf{TopiOCQA}} &\multicolumn{4}{c}{\textbf{QReCC}} \\\cmidrule{5-12}
        & & & &\textbf{MRR@3} &\textbf{NDCG@3} &\textbf{R@10} &\textbf{R@100} &\textbf{MRR@3} &\textbf{NDCG@3} &\textbf{R@10} &\textbf{R@100} \\\midrule
        \multirow{4}{*}{\textbf{Sparse(BM25)}} &Qwen2.5-3B &\xmark &\xmark &8.8 &7.5 &17.3 &36.1 &27.3 &25.0 &42.3 &64.6 \\
        &+ w/ SDPWU &\cmark &\xmark &14.8 &13.0 &27.8 &50.9 &47.6 &45.5 &64.3 &79.8 \\
        &+ w/ RIRS &\xmark &\cmark &32.6 &31.0 &56.2 &79.0 &51.7 &49.9 &71.9 &85.8 \\
        &+ ConvSearch-R1 &\cmark &\cmark &\textbf{35.2} &\textbf{33.5} &\textbf{57.8} &\textbf{79.9} &\textbf{56.5} &\textbf{54.8} &\textbf{76.3} &\textbf{88.1} \\
        \midrule
        \multirow{4}{*}{\textbf{Dense(ANCE)}} &Qwen2.5-3B &\xmark &\xmark &19.2 &18.3 &33.0 &46.7 &29.3 &27.2 &44.2 &59.0 \\
        &+ w/ SDPWU &\cmark &\xmark &22.2 &20.6 &38.7 &58.3 &39.9 &37.5 &59.3 &73.8 \\
        &+ w/ RIRS &\xmark &\cmark &49.2 &48.6 &70.4 &84.8 &47.7 &45.6 &68.3 &80.7 \\
        &+ ConvSearch-R1 &\cmark &\cmark &\textbf{51.4} &\textbf{51.3} &\textbf{72.0} &\textbf{85.7} &\textbf{49.7} &\textbf{47.7} &\textbf{69.8} &\textbf{81.6} \\
        \bottomrule
        \end{tabular}
    }
    \caption{Ablation results for both dense and sparse retrieval on the TopiOCQA and QReCC. \textbf{SDPWU} and \textbf{RIRS} denote the stage1 (\S~\ref{sec:SDPWU}) and stage2 (\S~\ref{sec:rirs}) of ConvSearch-R1, respectively.}
    \label{tab:ablation}
\end{table*}

For the evaluation of generalization ability, we trained models on the TopiOCQA training set and evaluated them on the QReCC test set, and vice versa. As shown in Table \ref{tab:generalization}, our method demonstrates superior generalization performance compared to other approaches. This significant improvement in generalization can be primarily attributed to our use of RL with a Rank-Incentive Reward. By employing a well-designed reward function, our training paradigm encourages the model to interact with a broader and more diverse set of high-quality data through trial and error, thereby enhancing its ability to generalize across different datasets.

\begin{figure*}[t]
    \centering
    \begin{subfigure}{0.48\textwidth}
        \centering
        \includegraphics[width=\linewidth]{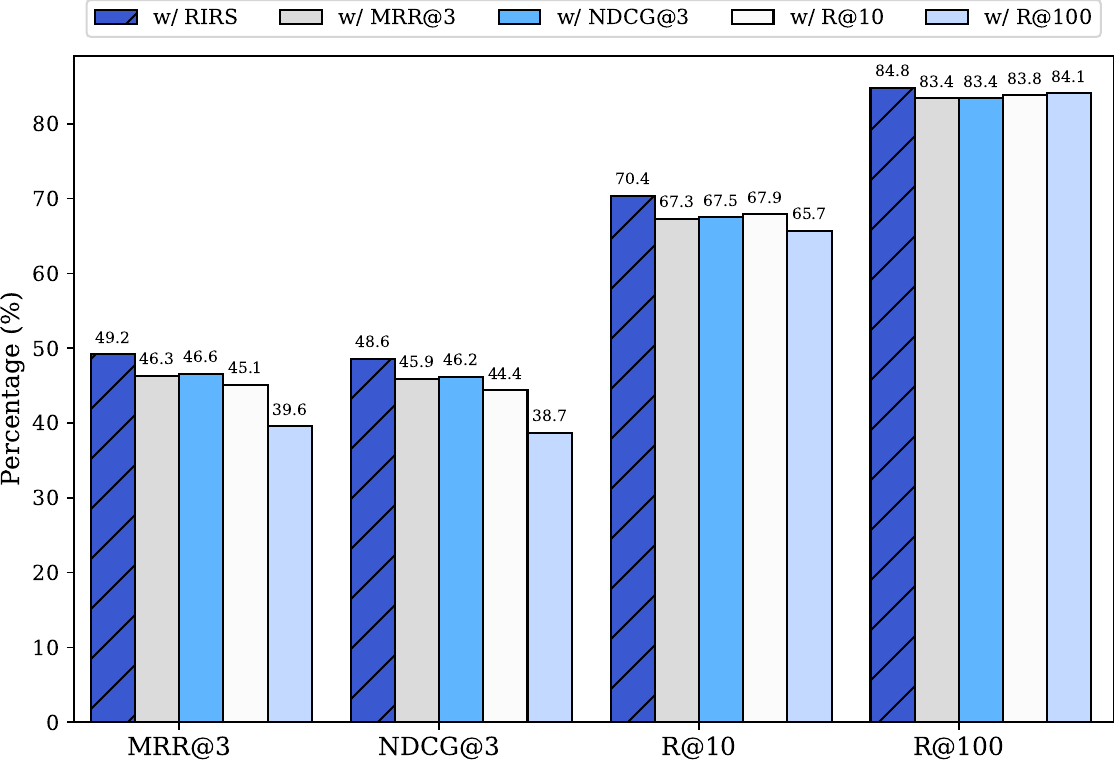}
        \caption{Comparison on TopiOCQA}
    \end{subfigure}
    \hfill
    \begin{subfigure}{0.48\textwidth}
        \centering
        \includegraphics[width=\linewidth]{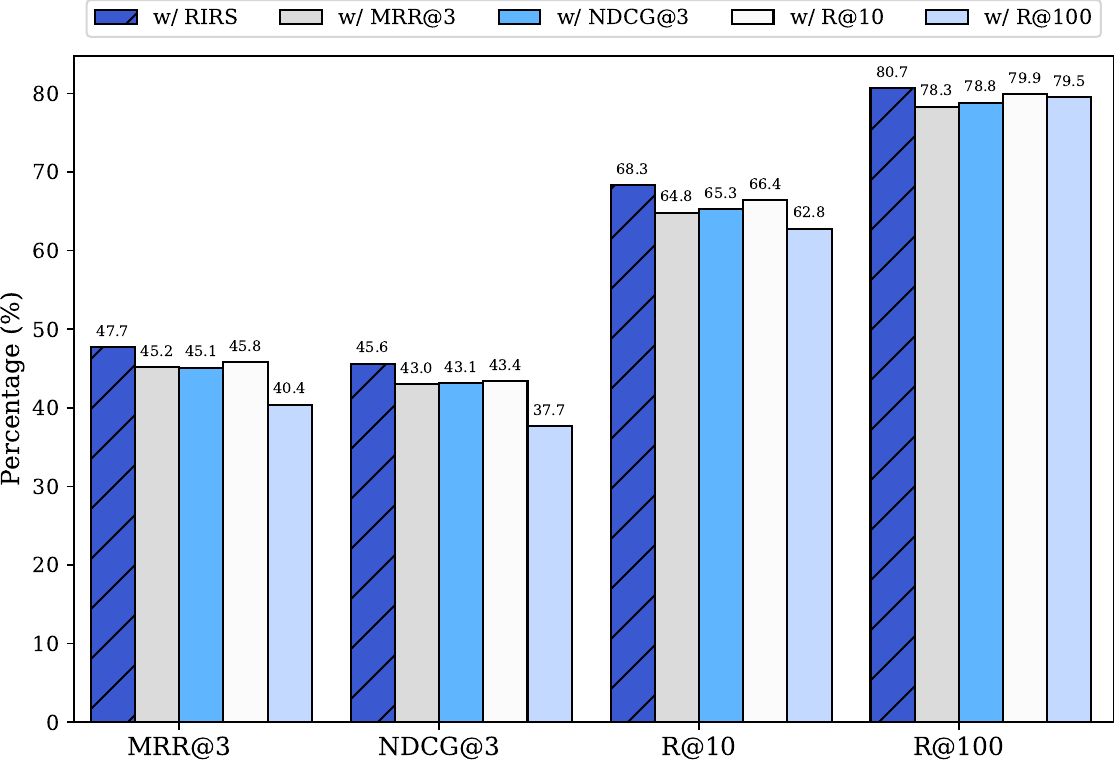}
        \caption{Comparison on QReCC}
    \end{subfigure}
    
    \caption{Comparison of different reward functions for dense retrieval on the TopiOCQA and QReCC datasets.}
    \label{fig:reward_ana}
\end{figure*}

\subsection{Scaling Behavior of Model Performance}
\label{sec:scaling_law}

To further validate the generalizability of our approach across models of varying parameter scales, we conducted experiments on the TopiOCQA dataset using the Qwen2.5 Series models, as shown in Figure \ref{fig:model_scale}. Our results demonstrate that our method consistently outperforms the SOTA baseline across almost all model sizes. Notably, as the model size increases, the performance gap between our method and the SOTA baseline becomes increasingly pronounced. Interestingly, even when using a relatively small 0.5B parameter model under dense (ANCE) retrieval, our approach still significantly surpasses the SOTA baseline. These findings indicate that ConvSearch-R1 exhibits strong generalizability across models of any scale.

\subsection{Ablation Study}
\label{sec:ablation_study}

\noindent \textbf{Overall Analysis.}\quad ConvSearch-R1 is a two-stage alignment framework. In Stage 1 (\S~\ref{sec:SDPWU}), an initial preference-aligned rewriter is obtained through self-distillation and preference-based filtering. In Stage 2 (\S~\ref{sec:rirs}), reinforcement learning with carefully designed rewards is employed to further align the rewriter with retriever preferences. To validate the effectiveness of each component, we conduct ablation studies. As shown in Table \ref{tab:ablation}, each component in ConvSearch-R1 plays a critical role across different retrievers and datasets, demonstrating the rationality and effectiveness of our framework’s design.

\smallskip

\noindent \textbf{Warm-Up Analysis.}\quad To validate the necessity of Stage 1 (\S~\ref{sec:SDPWU}), we selected checkpoints from various steps during the training process using different retrievers and on different datasets. These checkpoints were then evaluated accordingly. As shown in Figure \ref{fig:warmup_ana}, as the number of training steps increases, the MRR@3 score gradually improves and eventually reaches a plateau. At every training step, ConvSearch-R1 consistently outperforms the RIRS-only model (i.e., the model trained without Stage 1) in terms of MRR@3. These results provide strong evidence for the essential role of Stage 1 in enhancing model performance.

\smallskip

\noindent \textbf{Reward Analysis.}\quad To further validate the necessity of reward shaping in stage 2 (\S~\ref{sec:rirs}), we conducted comparative experiments on different reward functions using the Qwen2.5-3B model implemented with a dense retriever architecture. As shown in Figure \ref{fig:reward_ana}, the rewriter utilizing Rank-Incentive Reward Shaping consistently achieved the best performance across all settings. These results provide strong evidence for the necessity of reward shaping in our task.

\subsection{Case Study}
\label{sec:case_study}
\textbf{Why does ConvSearch-R1 lead to significant performance improvements?} To answer this question, we analyze a specific case selected from the QReCC test set, as shown in Appendix Table \ref{tab:case_study}. In this case, the user asks two questions within a single query, both involving coreference to previous context. In the early stages of training, the alignment between the rewriter and the retriever's preferences is limited. We observe that, during the reasoning process, the rewriter abandons the reformulation of the first question, resulting in a final rewrite that only addresses the second question. Consequently, the gold passage is not retrieved within the top 100 results. In the later stages of training, after extensive trial and error, the rewriter achieves a much higher degree of alignment with the retriever's preferences. We find that the rewriter repeatedly considers both user questions during reasoning and generates a rewrite that successfully resolves the coreference for both. Notably, the rewriter even generates pseudo-passages to supplement missing information. Thanks to this comprehensive consideration and the generation of pseudo-passages, the rewriter is able to retrieve the gold passage at the top-1 position for this case in the later training stage. These findings provide strong evidence that our method can effectively align with the retriever's preferences and achieve state-of-the-art performance on the benchmark.

\section{Conclusion}
In this paper, we propose ConvSearch-R1, a novel two-stage alignment framework for CQR that operates without the need for any external supervised data (reference rewrite). First, we generate high-quality SFT data via self-distillation and rank-based filtering. Then, we refine the model using RL with a Rank-Incentive Reward, aligning the rewriter with the retriever’s preferences through reasoning. To the best of our knowledge, we are the first to tackle the CQR task without relying on any form of external supervision. Experiments show state-of-the-art performance on in/out-of-domain datasets, consistent scaling with model size, and meaningful ablation insights.

\section*{Limitations}
While ConvSearch-R1 demonstrates promising results, several limitations remain that warrant further investigation. \textbf{(1) Limited Exploration of Larger Scales:} Due to resource constraints, we did not train extremely large models (e.g., 34B or 72B). Nonetheless, scaling experiments show consistent performance gains with increasing model size, suggesting effectiveness at larger scales. Given the deployment advantages of smaller models, we focused on models below 7B, where ConvSearch-R1 achieves superior results. \textbf{(2) Computational Overhead:} ConvSearch-R1 incorporates a reasoning-based paradigm, leading to longer output sequences and potentially higher inference cost. However, its 3B parameter size requires fewer resources than 7B-based methods, while still achieving strong performance.

\section*{Acknowledgments}
This work was supported by the National Natural Science Foundation of China (No. 62236004) and Shanghai Pilot Program for Basic Research - Fudan University 21TQ1400100 (22TQ018).

\bibliography{custom}

\appendix
\newpage
\section{ConvSearch-R1 VS Prior Works}
\label{appendix:diff}
\begin{figure*}
    \centering
    \includegraphics[width=\linewidth]{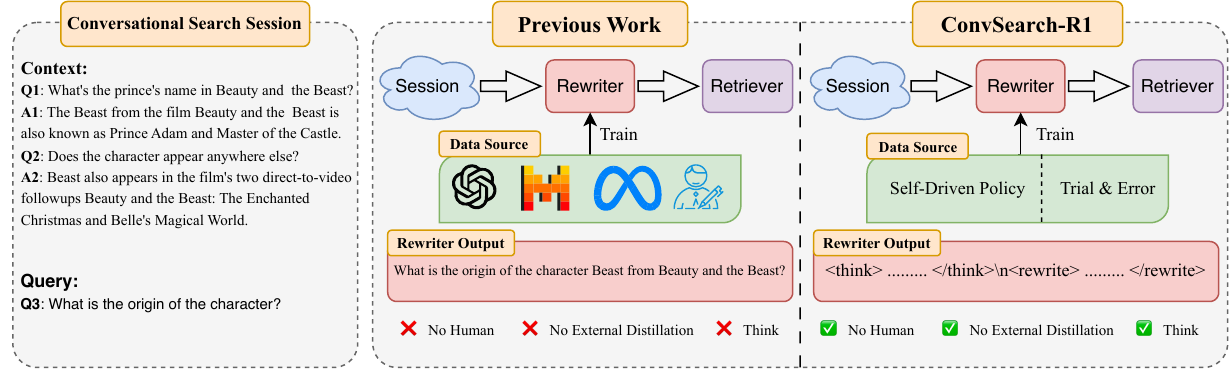}
    \caption{Comparison of ConvSearch-R1 with previous works.}
    \label{fig:diff}
\end{figure*}

As depicted in Figure \ref{fig:diff}, the distinction of ConvSearch-R1 from previous studies lies in the approach to obtaining rewrites. Prior research has always relied on external sources (e.g., human rewrites or powerful LLMs) to generate rewrites and has failed to endow the rewriter with reasoning capability. In contrast, ConvSearch-R1 acquires rewrites through self-distillation and RL, utilizing trial and error, and equips the rewriter with reasoning ability.

\section{Data Collection}
\label{appendix:data-collection}
We performed self-distillation on Qwen2.5-3B and Llama3.2-3B using the prompts specified in Figure \ref{fig:few-shot prompt} on the training set of TopiOCQA \citep{Adlakha2021TopiOCQAOC} and QReCC \citep{anantha-etal-2021-qrecc}. We retained only those samples that conformed to the required format and for which the rewritten query resulted in the gold passage being ranked first. The number of qualified samples for each model on the TopiOCQA and QReCC datasets is summarized in Table \ref{tab:self-driven-data-nums}.

\begin{table}[!htp]\centering
    \scriptsize
    \begin{tabular}{lcc}\toprule
    \multirow{2}{*}{\textbf{Model}} &\textbf{TopiOCQA} &\textbf{QReCC} \\\cmidrule{2-3}
    &\textbf{Sample Nums} &\textbf{Sample Nums} \\\midrule
    \textbf{Qwen2.5-3B} &5892 &7759 \\
    \textbf{Llama3.2-3B} &6034 &6623 \\
    \bottomrule
    \end{tabular}
    \caption{The number of samples obtained using the SDPWU (\S~\ref{sec:SDPWU}) on TopiOCQA and QReCC datasets for both Qwen2.5-3B and Llama3.2-3B models.}
    \label{tab:self-driven-data-nums}
\end{table}

\begin{figure}[!ht]
\begin{AIbox}{Prompt for few-shot learning:}

Given a query and its context, you must first think about the reasoning process in the mind to decontextualize the query by resolving coreference and omission issues. Then, provide the user with a rewrite that retains its original meaning and is as informative as possible to help search engines retrieve relevant documents effectively. The reasoning process and rewrite should be enclosed within <think> </think> and <rewrite> </rewrite> tags, respectively, i.e., <think> reasoning process here </think>\textbackslash n<rewrite> rewrite here </rewrite>.
        \\ \\
        Here is an example for your reference: \\
        \#\#\# Example Begin \#\#\# \\
        \{\textit{example}\} \\
        \#\#\# Example End \#\#\#
        \\ \\
        \#\#\# Context Begin \#\#\# \\
        \{\textit{context}\} \\
        \#\#\# Context End \#\#\#
        \\ \\
        Query: \{\textit{query}\}
\end{AIbox}
\caption{Prompt for few-shot learning.}\label{fig:few-shot prompt}
\end{figure}

\begin{figure}[!ht]
\begin{AIbox}{Prompt for training and inference:}

Given a query and its context, you must first think about the reasoning process in the mind to decontextualize the query by resolving coreference and omission issues. Then, provide the user with a rewrite that retains its original meaning and is as informative as possible to help search engines retrieve relevant documents effectively. The reasoning process and rewrite should be enclosed within <think> </think> and <rewrite> </rewrite> tags, respectively, i.e., <think> reasoning process here </think>\textbackslash n<rewrite> rewrite here </rewrite>.
        \\ \\
        \#\#\# Context Begin \#\#\# \\
        \{\textit{context}\} \\
        \#\#\# Context End \#\#\#
        \\ \\
        Query: \{\textit{query}\} \\
        Rewrite:
\end{AIbox}
\caption{Prompt for training and inference.}\label{fig:training prompt}
\end{figure}

\section{Results on Llama2-7B}
\label{appendix:res-on-llama2}
To ensure a fair comparison, we followed the experimental protocols of \citet{yoon-etal-2025-RetPO} (using Llama2-7B) and \citet{zhang-etal-2024-AdaQR} (using Mistral-7B), applying our method on the Llama2-7B\footnote{\url{https://huggingface.co/meta-llama/Llama-2-7b-chat-hf}} model under the same experimental settings. As shown in Table \ref{tab:res-on-llama2}, ConvSearch-R1 consistently achieves state-of-the-art performance on Llama2-7B. This result provides strong evidence for the effectiveness and generalizability of ConvSearch-R1 across different models. It is worth noting that Llama2-7B heavily relies on SFT to acquire robust format adherence capability; without SFT, the stability of RL training can be significantly compromised, resulting in worse performance.

\begin{table*}[!htp]\centering
    \scriptsize
    \begin{tabular}{lccccccccccc}\toprule
    \multirow{2}{*}{\textbf{Type}} &\multirow{2}{*}{\textbf{Method}} &\multicolumn{4}{c}{\textbf{TopiOCQA}} &\multicolumn{4}{c}{\textbf{QReCC}} &\multirow{2}{*}{\textbf{Avg}} \\\cmidrule{3-10}
    & &\textbf{MRR@3} &\textbf{NDCG@3} &\textbf{R@10} &\textbf{R@100} &\textbf{MRR@3} &\textbf{MRR@3} &\textbf{R@10} &\textbf{R@100} & \\\midrule
    \multirow{3}{*}{\textbf{Sparse(BM25)}} &Baseline SOTA &28.3 &26.5 &48.9 &73.1 &55.1 &52.5 &78.5 &93.7 &57.1 \\\cmidrule{2-11}
    &ConvSearch-R1 &35.1 &33.8 &58.0 &79.3 &55.7 &54.0 &76.8 &88.1 &\textbf{60.1} \\
    &w/o SFT &13.5 &11.9 &25.1 &48.4 &34.0 &31.5 &51.7 &74.4 &36.3 \\\midrule
    \multirow{3}{*}{\textbf{Dense(ANCE)}} &Baseline SOTA &38.5 &37.6 &61.6 &79.9 &47.2 &44.2 &70.7 &85.3 &58.1 \\\cmidrule{2-11}
    &ConvSearch-R1 &49.2 &49.0 &70.5 &84.8 &49.0 &47.1 &69.8 &82.3 &\textbf{62.7} \\
    &w/o SFT &23.7 &22.8 &38.1 &54.8 &33.6 &31.7 &50.6 &62.5 &39.7 \\
    \bottomrule
    \end{tabular}
    \caption{Results of ConvSearch-R1 using Llama2-7B as a backbone of the rewriter. For comparison, Baseline SOTA refers to the best performance achieved by any baseline method on each respective metric.}
    \label{tab:res-on-llama2}
\end{table*}

\begin{figure*}
    \centering
    \includegraphics[width=\linewidth]{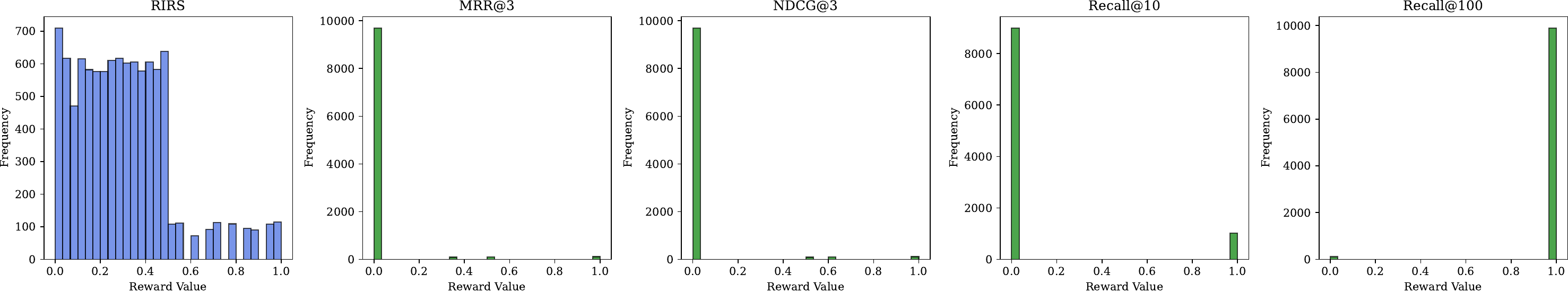}
    \caption{Comparison of Reward Sparsity.}
    \label{fig:reward_sparsity}
\end{figure*}

\section{Rank-Incentive Reward}
\label{appendix:rir}
To further evaluate the generalizability of the Rank-Incentive Reward across different types of functions, we designed experiments involving three distinct categories of functions. The formulations of these functions are as follows:

{\scriptsize
    \begin{gather}
        Reward(rank) = 
        \begin{cases} 
            a*rank + b, & \text{Piecewise Linear Func} \\
            e^{1-rank}, & \text{Exponential Decay Func} \\
            \frac{1}{rank} , & \text{Reciprocal Func}
        \end{cases}
    \end{gather}
}

\noindent The experimental results on Qwen3.2-3B are presented in Table \ref{tab:rir}. As shown, all methods achieve comparable performance, and each surpasses the baselines, reaching state-of-the-art results on TopiOCQA. In this paper, we use the Piecewise Linear Function as Rank-Incentive Reward to present the main results.

\begin{table*}[!th]
\centering
\small
        \begin{tabular}{lccccccc}\toprule
        \multirow{2}{*}{\textbf{Type}} &\multirow{2}{*}{\textbf{Method}} &\multicolumn{4}{c}{\textbf{TopiOCQA}} &\multirow{2}{*}{\textbf{Avg}} \\\cmidrule{3-6}
        & &\textbf{MRR@3} &\textbf{NDCG@3} &\textbf{R@10} &\textbf{R@100} & \\\midrule
        \multirow{4}{*}{\textbf{Sparse(BM25)}} &Baseline SOTA &28.3 &26.5 &48.9 &73.1 &44.2 \\\cmidrule{2-7}
        &Piecewise Linear Func &35.2 &33.5 &57.8 &79.9 &\textbf{51.6} \\
        &Exponential Decay Func &36.8 &35.5 &58.2 &79.6 &\textbf{52.5} \\
        &Reciprocal Func &34.8 &33.3 &57.6 &78.6 &\textbf{51.1} \\\midrule
        \multirow{4}{*}{\textbf{Dense(ANCE)}} &Baseline SOTA &38.5 &37.6 &61.6 &79.9 &54.4 \\\cmidrule{2-7}
        &Piecewise Linear Func &51.4 &51.3 &72.0 &85.7 &\textbf{65.1} \\
        &Exponential Decay Func &50.4 &50.0 &70.6 &85.5 &\textbf{64.1} \\
        &Reciprocal Func &50.6 &50.2 &72.0 &85.5 &\textbf{64.6} \\
        \bottomrule
        \end{tabular}
    \caption{Results of different Rank-Incentive Rewards. Baseline SOTA refers to the best performance achieved by any baseline method on each respective metric.}\label{tab:rir}
\end{table*}

\section{Experimental Details}
\label{appendix:exp_details}

\subsection{Datasets Details}
\label{appendix:data_details}

We use QReCC and TopiOCQA as our datasets in the experiments: (1) QReCC focuses on query rewriting. The overall task is relatively simple, and it provides human-rewritten queries. (2) TopiOCQA emphasizes topic shifts within conversations. It generally involves more conversation turns than QReCC and poses a higher level of difficulty. However, it does not provide human-rewritten queries.

For all datasets, we remove samples that lack gold passages. In the case of QReCC, some samples have gold passages but no corresponding answers for the queries. Following TopiOCQA, we assign \texttt{UNANSWERABLE} as the answer for such queries. For detailed dataset statistics, please refer to Table \ref{tab:dataset_details}. All datasets used in this paper are supported for academic research.

\begin{table}[!htp]\centering
    \scriptsize
    \begin{tabular}{lrrrrr}\toprule
    Dataset &Split &\#Conv. &\#Turns(Qry.) &\#Collection \\\midrule
    \multirow{2}{*}{TopiOCQA} &train &3,509 &45,450 &\multirow{2}{*}{25M} \\
    &test &205 &2,514 & \\\midrule
    \multirow{2}{*}{QReCC} &train &10,823 &29,596 &\multirow{2}{*}{54M} \\
    &test &2,775 &8,124 & \\
    \bottomrule
    \end{tabular}
    \caption{Statistics of conversational search datasets.}\label{tab:dataset_details}
\end{table}

\subsection{Metrics Details}
\label{appendix:metric_details}

In this study, we employ three widely used metrics to evaluate the performance of our method: Mean Reciprocal Rank at K (MRR@K), Normalized Discounted Cumulative Gain at K (NDCG@K), and Recall at K (Recall@K). These metrics provide complementary perspectives on ranking quality and retrieval effectiveness.

\textbf{MRR@K} measures the average reciprocal rank of the first relevant item within the top K results across all queries. It emphasizes the importance of retrieving a relevant item as high as possible in the result list, rewarding systems that return relevant results earlier.

\textbf{NDCG@K} evaluates the ranking quality by considering both the position and the graded relevance of items within the top K results. This metric assigns higher importance to relevant items appearing higher in the ranking and accounts for scenarios where relevance is not binary, thus providing a more nuanced assessment of ranking effectiveness.

\textbf{Recall@K} quantifies the proportion of relevant items that are successfully retrieved within the top K results. It reflects the system's ability to cover as many relevant items as possible in the truncated result list, providing insight into the coverage of relevant content.

\subsection{Baselines Details}
\label{appendix:baselines_details}

In our experiments, we compare the following baselines: 
(1) \textbf{Human Rewrite}: Manually annotated query rewrites; 
(2) \textbf{Raw}: The user’s original query within the dialogue context; 
(3) \textbf{DS-R1-Distill-Qwen-7B}: A reasoning model distilled from DeepSeek-R1; 
(4) \textbf{QuReTeC}~\citep{Voskarides2020QueryRF}: A bidirectional transformer model for resolving underspecified queries in multi-turn conversational search, using distant supervision for training data generation; (5) \textbf{T5QR}~\citep{Lin2020ConversationalQR}: A sequence-to-sequence model based on T5, fine-tuned for conversational question reformulation; (6) \textbf{CONQRR}~\citep{Wu2021CONQRRCQ}: A reinforcement learning-based approach that rewrites conversational queries to optimize retrieval with any retriever; (7) \textbf{ConvGQR}~\citep{mo-etal-2023-convgqr}: Combines query rewriting and expansion using generative pre-trained models, incorporating knowledge infusion for better retrieval; (8) \textbf{EDIRCS}~\citep{mao-etal-2023-search}: A non-autoregressive, text-editing model that selects most rewrite tokens from dialogue context, trained with search-oriented objectives for efficient query reformulation; (9) \textbf{IterCQR}~\citep{jang-etal-2024-itercqr}: Iteratively optimizes query reformulation based on information retrieval signals without human supervision; (10) \textbf{ADACQR}~\citep{lai-etal-2025-adacqr}: Aligns reformulation models with both sparse and dense retrievers via a two-stage training strategy; (11) \textbf{CHIQ-Fusion}~\citep{mo-etal-2024-CHIQ}: Improves conversational history quality with open-source LLMs before query generation; (12) \textbf{RETPO}~\citep{yoon-etal-2025-RetPO}: Fine-tunes a language model using large-scale retriever feedback to generate rewrites aligned with retrieval preferences; (13) \textbf{AdaQR}~\citep{zhang-etal-2024-AdaQR}: Trains query rewriting models with limited annotations by using conversational answer probability as a reward, eliminating the need for passage labels; (14) \textbf{LLM4CS}~\citep{mao-etal-2023-LLM4CS}: Uses LLMs to generate and aggregate multiple query rewrites and hypothetical responses, robustly representing user intent; (15) 
\textbf{InfoCQR}~\citep{Ye2023EnhancingCS}: Utilizes LLMs as query rewriters and editors, then distills their capabilities into smaller models for efficiency; (16) \textbf{Qwen2.5-3B}: A lightweight multi-modal AI model from Alibaba Cloud, delivering strong performance with fewer parameters; (17) \textbf{Llama2.5-3B}: A compact version of Meta’s Llama, optimized for efficient, edge-device processing with robust text understanding and generation.

\begin{table*}[!th]
    \resizebox{\textwidth}{!}{
        \begin{tabular}{llccccccccc}\toprule
        \multirow{2}{*}{\textbf{Retriever}} &\multirow{2}{*}{\textbf{Method}} &\multicolumn{4}{c}{\textbf{TopiOCQA}} &\multicolumn{4}{c}{\textbf{QReCC}} \\\cmidrule{3-10}
        & &\textbf{MRR@3} &\textbf{NDCG@3} &\textbf{R@10} &\textbf{R@100} &\textbf{MRR@3} &\textbf{NDCG@3} &\textbf{R@10} &\textbf{R@100} \\\midrule
        ANCE &Llama3.2-3B &12.8 &11.7 &22.6 &38.1 &19.1 &17.4 &30.4 &44.1 \\
        ANCE & + ConvSearch-R1 (ours) &50.5 &50.1 &72.0 &86.3 &50.2 &48.1 &70.6 &82.8 \\
        RepLLaMA &Llama3.2-3B &13.1 &14.5 &23.1 &39.9 &21.1 &19.9 &32.4 &45.7 \\
        RepLLaMA & + ConvSearch-R1 (ours) &51.8 &52.4 &73.4 &88.5 &52.4 &49.3 &71.6 &82.9 \\\midrule
        ANCE &Qwen2.5-3B &19.2 &18.3 &33.0 &46.7 &29.3 &27.2 &44.2 &59.0 \\
        ANCE & + ConvSearch-R1 (ours) &51.4 &51.3 &72.0 &85.7 &49.7 &47.7 &69.8 &81.6 \\
        RepLLaMA &Qwen2.5-3B &21.1 &20.3 &34.9 &47.9 &30.3 &29.5 &46.6 &59.9 \\
        RepLLaMA & + ConvSearch-R1 (ours) &53.6 &53.4 &73.5 &85.0 &51.7 &49.7 &70.9 &81.1 \\
        \bottomrule
        \end{tabular}
    }
    \caption{Performance comparison of our proposed method ConvSearch-R1 when integrated with different underlying retrievers, demonstrating its robustness and generalizability.}
    \label{tab:diff_retriever}
\end{table*}

\begin{table*}[!th]
    \resizebox{\textwidth}{!}{
        \begin{tabular}{lcccccccccccc}\toprule
        \multirow{2}{*}{\textbf{Method}} &\multicolumn{4}{c}{\textbf{TREC CAsT 2019}} &\multicolumn{4}{c}{\textbf{TREC CAsT 2020}} &\multicolumn{3}{c}{\textbf{TREC CAsT 2021}} \\\cmidrule{2-12}
        &\textbf{MRR@3} &\textbf{NDCG@3} &\textbf{R@10} &\textbf{R@100} &\textbf{MRR@3} &\textbf{NDCG@3} &\textbf{R@10} &\textbf{R@100} &\textbf{MRR@3} &\textbf{NDCG@3} &\textbf{R@10} \\\midrule
        Human Rewrite &74.0 &46.1 &- &38.1 &59.1 &42.2 &- &\textbf{46.5} &- &- &- \\
        QuReTeC &68.9 &43.0 &- &33.7 &43.0 &28.7 &- &34.6 &- &- &- \\
        T5QR &70.1 &41.7 &- &33.2 &42.3 &29.9 &- &35.3 &- &- &- \\
        EDIRCS &- &44.0 &- &35.5 &- &30.8 &- &37.5 &- &- &- \\
        ConvGQR &70.8 &43.4 &- &- &46.5 &33.1 &- &- &- &- &- \\
        RepLLaMA &62.4 &31.6 &10.6 &- &26.8 &18.3 &10.4 &- &47.4 &32.7 &19.6 \\
        E5-Mistral &62.2 &31.3 &9.5 &- &22.0 &15.4 &8.4 &- &48.2 &32.5 &17.3 \\
        LLM-Embedder &63.3 &36.6 &11.4 &- &25.2 &15.4 &8.7 &- &46.8 &31.2 & \\
        HyDE &55.6 &39.2 &10.0 &- &44.8 &29.3 &16.9 &- &- &- &- \\
        Query2doc &58.8 &42.4 &11.6 &- &48.6 &32.5 &17.3 &- &- &- &- \\
        InstructorR &61.2 &46.6 &10.4 &- &43.7 &29.6 &8.3 &- &46.7 &32.5 &18.4 \\
        LLM4CS &70.4 &46.8 &11.7 &- &58.6 &41.5 &\textbf{19.3} &- &66.1 &46.9 &24.4 \\
        CHIQ-Fusion &73.3 &50.5 &12.9 &- &54.0 &38.0 &\textbf{19.3} &- &62.9 &46.5 &25.2 \\\midrule
        ConvSearch-R1 (ours) &\textbf{76.6} &\textbf{54.0} &\textbf{14.2} &\textbf{42.0} &\textbf{70.6} &\textbf{42.6} &15.6 &43.8 &\textbf{73.2} &\textbf{52.1} &\textbf{26.2} \\
        \bottomrule
        \end{tabular}
    }
    \caption{Generalization performance of ConvSearch-R1 and baselines on the TREC CAsT 2019, 2020, and 2021 datasets using ANCE as a retriever.}
    \label{tab:further_generalization}
\end{table*}

\subsection{Implementation Details}
\label{appendix:imple_details}

All experiments are conducted based on the verl~\citep{Sheng2024verl}. The training process consists of two main stages: SFT and RL. During the SFT stage, we apply the following hyperparameters across all experiments: a batch size of 64, a maximum sequence length of 3072, 2 training epochs, and a learning rate of 1e-5. For the RL stage, the hyperparameters are set as follows for all experiments: a batch size of 128, a maximum prompt length of 1536, a maximum response length of 1024, a learning rate of 1e-6, 100 learning rate warmup steps, a clipping ratio of 0.2, a KL loss coefficient of 0.001, and a rollout sample size of \textit{n}=8 with a sampling temperature of 0.7. The number of training epochs is set to 6 for TopiOCQA and 9 for QReCC, respectively.

For evaluation, we follow the protocol described in \citet{zhang-etal-2024-AdaQR}. The BM25 parameters are set to \(k_1\) = 0.9 and \textit{b} = 0.4 for TopiOCQA, and \(k_1\) = 0.82 and \textit{b} = 0.68 for QReCC. The lengths of the query, concatenated input, and passage are truncated to 64, 512, and 384 tokens, respectively. 

In this paper, we use the Instruct-tuned versions of Llama3.2-3B\footnote{\url{https://huggingface.co/meta-llama/Llama-3.2-3B-Instruct}} and Qwen2.5-3B\footnote{\url{https://huggingface.co/Qwen/Qwen2.5-3B-Instruct}} for all experiments.

\section{Generalization}
\label{appendix:generalization}

\subsection{Retrievers}
\label{appendix:diff_retriever}

For a fair comparison against the baselines in our main experiments, we employed BM25 and ANCE as retrievers. Recognizing that these earlier models may not reflect contemporary application scenarios, we conducted supplementary experiments with a more recent dense retriever, RepLLaMA \citep{Ma2023repLLaMA}. As shown in Table \ref{tab:diff_retriever}, the adoption of a more powerful retriever yields further performance gains. This result further demonstrates the robustness and generalizability of ConvSearch-R1.

\subsection{TREC CAsT}
\label{appendix:trec_cast}

To rigorously evaluate the generalization capabilities of our proposed method, ConvSearch-R1, we conducted extensive experiments on a series of standard conversational search benchmarks: TREC CAsT 2019 \citep{Dalton2020trec2019}, 2020 \citep{Dalton2020trec2020}, and 2021 \citep{Dalton@2021}. This multi-year evaluation allows us to assess how well ConvSearch-R1 performs on datasets with varying characteristics and complexities, which it was not explicitly trained on. Table \ref{tab:further_generalization} presents the comparative results of ConvSearch-R1 against a comprehensive suite of baseline methods, using ANCE as the underlying dense retriever for all models to ensure a fair comparison. Its capacity to consistently outperform a diverse and competitive set of baselines, including human-level rewriting and state-of-the-art LLM methods, solidifies its position as a robust and highly effective solution for conversational query rewriting.

\begin{table*}
\centering
\small
        \begin{tabular}{lccccccccc}\toprule
        \multirow{2}{*}{\textbf{Method}} &\multicolumn{4}{c}{\textbf{TopiOCQA}} &\multicolumn{4}{c}{\textbf{QReCC}} \\\cmidrule{2-9}
        &\textbf{Num} &\textbf{ATPS} &\textbf{Time} &\textbf{TPS} &\textbf{Num} &\textbf{ATPS} &\textbf{Time} &\textbf{TPS} \\\midrule
        DS-R1-Distill-Qwen-7B &2514 &435 &194 &0.077 &8209 &418 &571 &0.07 \\
        RETPO (7B) &2514 &108 &113 &0.045 &8209 &107 &363 &0.044 \\
        Qwen2.5-3B (CoT) &2514 &127 &52 &0.021 &8209 &131 &176 &0.021 \\
        ConvSearch-R1 (3B) &2514 &386 &123 &0.049 &8209 &387 &435 &0.053 \\
        \bottomrule
        \end{tabular}
    \caption{Latency analysis between ConvSearch-R1 and baselines. The time is recorded in seconds.}
    \label{tab:efficiency}
\end{table*}

\section{Latency Analysis}
\label{appendix:latency}

Under the same settings, based on the vLLM \citep{kwon2023vllm} framework, we obtain the latency analysis shown in the table \ref{tab:efficiency}. Here, TPS refers to consumption time per sample, and ATPS denotes average token per sample. It can be seen that although ConvSearch-R1 increases the output length due to reasoning, its TPS is close to that of RETPO, and significantly shorter than that of DS-R1-Distill-Qwen-7B. At the same time, compared to Qwen2.5-3B, its TPS remains within an acceptable range. All of this demonstrates the efficiency of ConvSearch-R1.

\begin{table*}[!htp]\centering
\scriptsize
\begin{tabular}{p{0.95\textwidth}}
\toprule
\textbf{Conversation} \\\midrule
\#\#\# Context Begin \#\#\# \\
Q1: What was Wallis' relationship with Edward? \\
A1: In January 1934, while Lady Furness was away in New York City, Wallis Simpson allegedly became Prince Edward's mistress. \\
Q2: Were they romantically involved? \\
A2: Edward denied this to his father, despite his staff seeing them in bed together as well as evidence of a physical sexual act. \\
Q3: When did they marry? \\
A3: Edward VIII married Wallis Simpson in France on 3 June 1937, after her second divorce became final. Later that year, the couple toured Germany. \\
Q4: How did Wallis and Earl meet? \\
A4: In April 1916, Wallis Simpson met Earl Winfield Spencer Jr., a U.S. Navy aviator, at Pensacola, Florida, while visiting her cousin Corinne Mustin. \\
Q5: Did Earl have to leave due to World War I? \\
A5: After the United States entered the First World War in 1917, Earl Spencer was posted to San Diego. \\
Q6: Where did Earl and Wallis live during this time? \\
A6: Earl Spencer was posted to San Diego; they remained there until 1921. \\
Q7: Was their marriage happy? \\
A7: By September 1925, Wallis Simpson and her husband Earl Spencer were living apart. \\
\#\#\# Context End \#\#\# \\ \\
Query: Did they stay together or did they eventually get divorced? \\
\midrule
\textbf{ConvSearch-R1 (Initial Stage)} \\\midrule
\textcolor{blue}{<think>} According to A7, Wallis Simpson and Earl Spencer were living apart by September 1925. While the context does not explicitly state that they divorced, their separation suggests the end of their marriage. Since the original question asks whether they stayed together or eventually divorced, and we know they separated by 1925, it is reasonable to refine the question based on available information. \textcolor{blue}{</think>} \\
\textcolor{teal}{<rewrite>} Did Wallis Simpson and Earl Spencer eventually get divorced? \textcolor{teal}{</rewrite>} \colorbox[HTML]{ffe5ec}{(\textit{Not Found})} \\\midrule
\textbf{ConvSearch-R1 (Final Stage)} \\\midrule
\textcolor{blue}{<think>} The user is asking about the relationship and eventual outcome of Wallis Simpson and Earl Spencer's marriage. The previous context mentions their first meeting and that Wallis visited her cousin in Pensacola, Florida, in 1916. It also notes that Earl Spencer was a U.S. Navy aviator and their marriage was unhappy. Wallis and Earl's marriage eventually led to a divorce as Wallis had her second divorce by 1936, when she became involved with Edward VIII. The question seeks information about whether they were together at all or if they ultimately divorced. Context mentions their marriage was unhappy and they lived apart by 1925. There are no direct statements about getting divorced, but their separation led to their eventual divorce.\textcolor{blue}{</think>}\\
\textcolor{teal}{<rewrite>} Did Wallis Simpson and Earl Spencer's marriage last? Wallis Simpson was first introduced to Earl Spencer Jr. in 1916 when she visited her cousin in Pensacola, Florida. After the United States entered World War I in 1917, Earl Spencer was posted to San Diego. Wallis and Earl were married in 1928, but their marriage was described as unhappy. By 1925, Wallis and her husband separated and by 1936, they were living apart. Wallis had her second divorce in 1936, leading to her marriage to Prince Edward VIII in 1937. Did Wallis and Earl's marriage eventually end in divorce? Wallis and Earl Spencer's marriage was unhappy and they eventually separated and lived apart. Their first marriage was dissolved by 1936, with Wallis taking her second divorce. She and Earl did not stay together. Later, during her relationship with Prince Edward VIII, Wallis and Earl's whereabouts were not known publicly. They were living apart by 1935-1936, which led to their divorce. Did their marriage end with a divorce, and how did their relationship with each other and with each other's families fare during this period? In 1921, the couple moved to England, but eventually, their marriage broke down. Wallis and Earl's marriage was not happy, and they lived separately for a long time. After their divorce in 1928, they maintained a cordial relationship, but their marriage was indeed marked by separation and eventually divorce. After their divorce, Earl moved to the United States and later returned to England, where he died in 1954. Wallis Simpson went on to marry Prince Edward VIII in 1937, and the couple toured Germany in the same year.\textcolor{teal}{</rewrite>}  \colorbox[HTML]{ffe5ec}{(\textit{rank 1})} \\
\bottomrule
\end{tabular}
\caption{Case study on a example from QReCC using Qwen2.5-3B as a backbone.}\label{tab:case_study}
\end{table*}

\end{document}